%% file: main.tex
\pdfoutput=1

\documentclass[11pt]{article}

\usepackage[final]{acl}

\usepackage{times}
\usepackage{latexsym}

\usepackage[T1]{fontenc}

\usepackage[utf8]{inputenc}

\usepackage{microtype}

\usepackage{inconsolata}

\usepackage{graphicx}
\usepackage{booktabs}
\usepackage{inconsolata}
\usepackage{import}
\usepackage{microtype}
\usepackage{layout}
\usepackage{tabularx, makecell}
\usepackage{booktabs}
\usepackage{mathrsfs}
\usepackage{amssymb} 
\usepackage{url}
\usepackage{graphicx}
\usepackage{xspace,paralist}
\usepackage{times,latexsym}
\usepackage{amsmath}
\usepackage{appendix}
\usepackage{comment} 
\usepackage{enumitem}
\usepackage{makecell}
\usepackage{multirow}
\usepackage{xcolor}
\usepackage{graphicx}
\usepackage{cleveref}
\usepackage{tcolorbox}
\usepackage{adjustbox}
\usepackage{relsize}
\usepackage[normalem]{ulem}

\newcommand{\model}[1]{\texttt{#1}\xspace}

\newcommand{\gptfouro}{\model{GPT-4o}}

\newcommand{\whisperbase}{\model{Whisper-base}}
\newcommand{\whisperlarge}{\model{Whisper-large-v3}}

\definecolor{optimalgreen}{HTML}{2b8a3e}
\useunder{\uline}{\ul}{}

\title{Exploring the Potential of Multimodal LLM with\\ Knowledge-Intensive Multimodal ASR}

\author{Minghan Wang\textsuperscript{1}, Yuxia Wang\textsuperscript{2}, Thuy-Trang Vu\textsuperscript{1},\\ \textbf{Ehsan Shareghi}\textsuperscript{1}, \textbf{Gholamreza Haffari}\textsuperscript{1} \\
  \textsuperscript{1}Department of Data Science \& AI, Monash University \\ \textsuperscript{2}MBZUAI \\
  \texttt{\{minghan.wang,trang.vu1,ehsan.shareghi,gholamreza.haffari\}@monash.edu}\\
  \texttt{yuxia.wang@mbzuai.ac.ae}
}

\begin{document}
\maketitle
\input{sections/0_abstract}

\input{sections/1_intro}

\input{sections/3_method}

\input{sections/4_experiment}
\input{sections/5_eval_protocol}

\input{sections/6_analysis}
\input{sections/7_conclusion}
\input{sections/limitation_ethical}
\input{sections/9_acknowledgment}

\bibliography{custom}

\input{sections/8_appendix}

\end{document}

%% file: sections/0_abstract.tex
\begin{abstract}


Recent advancements in multimodal large language models (MLLMs) have made significant progress in integrating information across various modalities, yet real-world applications in educational and scientific domains remain challenging. 
This paper introduces the Multimodal Scientific ASR (MS-ASR) task, which focuses on transcribing scientific conference videos by leveraging visual information from slides to enhance the accuracy of technical terminologies.
Realized that traditional metrics like WER fall short in assessing performance accurately, prompting the proposal of severity-aware WER (SWER) that considers the content type and severity of ASR errors. 
We propose the Scientific Vision Augmented ASR (SciVASR) framework as a baseline method, enabling MLLMs to improve transcript quality through post-editing. 
Evaluations of state-of-the-art MLLMs, including GPT-4o, show a 45\% improvement over speech-only baselines, highlighting the importance of multimodal information integration.\footnote{Our code is available at \url{https://github.com/yuriak/MS-ASR}}

\end{abstract}

%% file: sections/1_intro.tex
\section{Introduction}

Recent years have witnessed remarkable progress in multimodal large language models (MLLMs), with GPT-4o achieving impressive performance on tasks involving the integration of information from multiple data modalities such as text, images, speech and video \citep{openai2024gpt4}. However, many real-world applications, particularly in the educational and scientific domains, present significant challenges that push the boundaries of current multimodal capabilities~\citep{gupta2023unsupervised,benedetto2024abstractive}. 

\begin{figure}
    \centering
    \resizebox{1\columnwidth}{!}{%
        \includegraphics{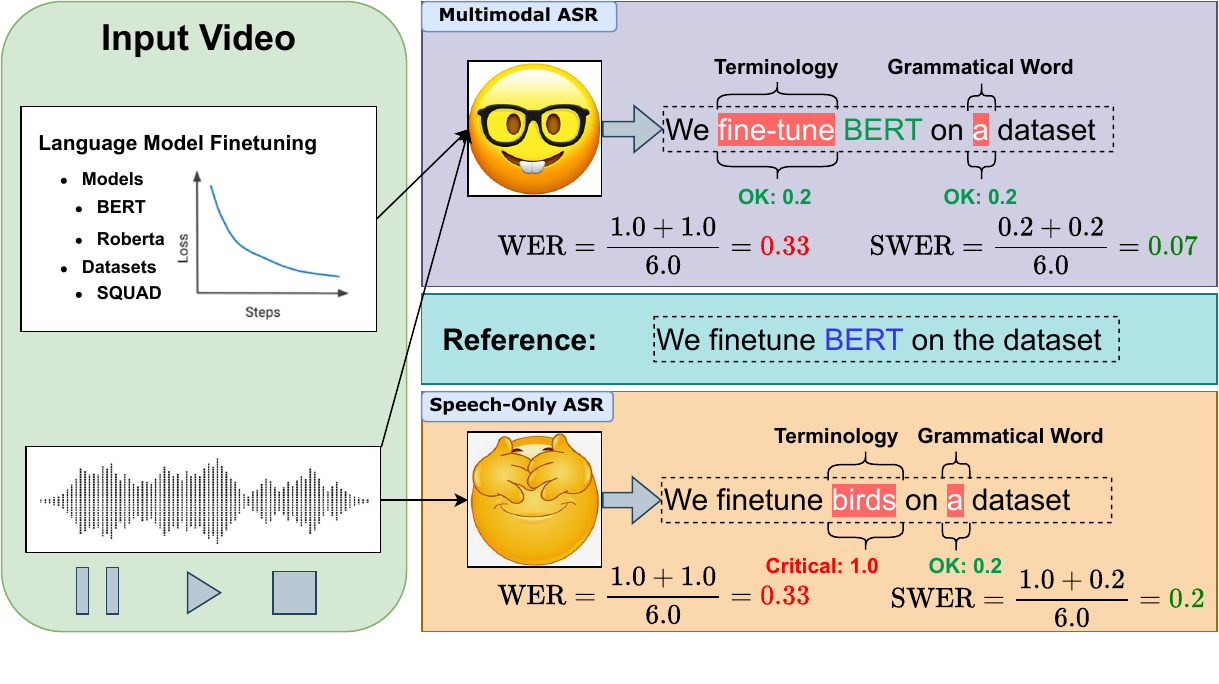}
    }
    \caption{
    This figure illustrates the importance of visual information in accurately recognizing the terminology \textit{BERT}. It also introduces our proposed evaluation metrics Severity-aware WER (SWER) by calibrating WER with the severity of ASR errors.
    }
    \label{fig:visual_context_importance}
\end{figure}

One such challenging task is multimodal automatic speech recognition (ASR) from knowledge-intensive presentation videos, which
involves not only transcribing spoken content but also understanding and integrating visual information from images or videos \citep{oneata2022improving, chang23c_interspeech}.
Presentation videos, often found in scientific conferences or educational contexts, contain dense, knowledge-rich content spanning various topics and concepts. Effective recognition and comprehension of such content are crucial for applications such as automatic transcript generation \citep{angrave2020benefits}, simultaneous translation~\citep{wang2024simultaneousmachinetranslationlarge,wang2024conversationalsimulmtefficientsimultaneous} in conference venue, lecture summarization~\citep{benedetto2024abstractive} and accessibility tools for researchers and learners \citep{mccarron2021creating}.
ASR on such videos presents several challenges, including keeping pace with the rapid evolution of technical domain knowledge and handling the variability in speaker accents, proficiency in English, and speech fluency~\citep{Cettolo2004}.
Mitigating these challenges requires ASR models to effectively extract and incorporate information from the speech and presentation slides in the video recordings (see \Cref{fig:visual_context_importance}).

In this paper, we investigate whether the current multimodal systems can successfully generate transcripts for presentation videos. We propose the task of transcribing scientific conference presentation recordings, namely Multimodal Scientific ASR (MS-ASR).
Unlike conventional multimodal ASR tasks that focus on motion-rich scenarios, MS-ASR focuses on knowledge-intensive content, requiring accurate transcription of domain-specific terminology using visual information. 
Standard evaluation metrics like word error rate (WER) fail to accurately assess these specialized terms, leading to biased performance evaluations. To address this, we propose a more nuanced evaluation protocol: severity-aware WER (SWER) for the task, which calibrates WER based on the severity of ASR errors, aligning system performance more closely with human evaluation.




Solving the proposed MS-ASR task requires proficiency in several areas, including speech recognition, visual information extraction and integrating them to post-edit the transcript. To this end, we propose a zero-shot inference framework named Scientific Vision Augmented ASR (SciVASR) that leverages MLLMs to collaboratively produce contextually accurate transcripts. We further explore the potential of the current SoTA MLLM, such as GPT-4o, for end-to-end post-editing the transcript.
Experimental results demonstrate that MLLMs can effectively improve the quality of the transcript compared to the speech-only ASR baseline, where visual information plays a key role for all MLLMs including GPT-4o to obtain the improvements.
We found that speech-only ASR suffers from high term errors (298) and 43.04\% critical errors, adding visual information not only reduces 152 term errors but also 28\% critical errors (Section \ref{sec:reevaluate}).
Human evaluation confirmed that remaining errors are negligible to affect the understanding on transcripts.

The contributions of this paper are threefold.
\begin{itemize}
    \item We introduce the Multimodal Scientific ASR (MS-ASR) task and evaluate both open- and proprietary SoTA MLLMs with our proposed zero-shot inference framework Sci-VASR. 
    \item We propose a more fine-grained ASR evaluation framework and severity-aware WER (SWER) metric. It can systematically assess transcription accuracy across multiple dimensions, closely reflecting human evaluation.
    \item Our experiments demonstrate that the proposed task effectively evaluates MLLMs' multimodal capabilities, with GPT-4o performing the best, achieving a 45\% improvement in SWER over the speech-only baseline.
\end{itemize}

%% file: sections/3_method.tex
\input{tables/dataset-stats}
\section{Multimodal Scientific ASR}
\label{sec:task}
\subsection{Task Formulation}
Given a presentation video consisting of a sequence of visual frames and a corresponding speech audio sequence, the goal is to generate an accurate transcript of the spoken content in the video. The visual frames typically contain presentation slides with text, images, diagrams and other visual aids relevant to the presented topic. The speech audio comprises the spoken explanations and narrations from the presenter, which are closely coupled with the visual content displayed on the slides (\Cref{fig:visual_context_importance}). The knowledge-intensive and technical nature of the presentation content requires the ASR model the ability to recognize domain-specific terminologies and effectively integrate information from both visual and speech modalities. 
Unlike conventional ASR task, which operates at sentence-level utterances over well-split speech segments, we treat each video as a whole and generate a presentation-level transcript for the entire video.
This task formulation aims to evaluate the accuracy of the ASR system on long videos and also aligns better with the real-world application where pre-segmented utterances are not typically available.

Standard ASR evaluation metrics, such as WER, can be used to evaluate model performance on this task. In our preliminary experiments, we found that WER does not align well with human evaluations and fails to adequately measure the model's ability to recognize technical terminologies (\S\ref{sec:preliminary_result}). Therefore, we propose severity-aware evaluation protocol that calibrates WER with the severity of ASR errors, providing a more nuanced and accurate assessment of model performance (\S\ref{sec:evaluationframework}).




\subsection{ACL60/60 Dataset}

We utilize the ACL 60/60 dataset comprising video recordings of accepted papers at ACL 2022~\citep{elizabeth2023acl6060}.
Each recording lasts 10-15 minutes, with a speaker presenting their paper in English.
It was primarily used for speech translation evaluation but contains high-quality human-annotated ASR transcripts and terminologies, making it a suitable dataset for our research.

This dataset comprises 10 video recordings, in total of around 120 minutes, with 5 videos split as the development set and 5 as the test set. 
These videos are uniformly sampled based on the speaker's nationality, accent, gender, English proficiency, and topic, as shown in Table \ref{tab:dataset_meta_info}. 
We also estimate the difficulty of each video by calculating the ratio of the conditional perplexity of the baseline speech generated by the text-to-speech system and the real speech. The detailed calculation of the difficulty estimation is in \Cref{sec:difficuty-score}.


Compared to other multimodal ASR datasets like How2~\citep{sanabria2018how2}, the ACL60/60 dataset differs due to its greater sample-wise length (presentation level vs utterance level), more specific domain coverage (scientific domain in NLP vs daily life), and higher knowledge density, making it an ideal benchmark for the proposed task.



\section{Scientific Vision-Augmented ASR}
\label{sec:msr}

Due to the nature of the long, knowledge-rich, and un-segmented videos, to effectively tackle the task of transcribing presentation videos, a multimodal ASR must excel at four subtasks: (i) \emph{speech recognition} to accurately transcribe the speech audio, (ii) \emph{video segmentation} to identify slide boundaries, (iii) \emph{visual information extraction} from each slide, and (iv) integrating the visual information to \emph{post-edit} the initial speech recognition transcript. In this section, we present our approach to address these subtasks, illustrated in \Cref{fig:architecture}.



\subsection{Speech Recognition}
We utilize state-of-the-art ASR models, such as \whisperbase~\citep{radford2023whisper}
to transcribe the video's speech into an initial transcript along with corresponding timestamps.
While this provides a reasonable starting point, the transcript may contain errors, especially in technical terminology, due to its knowledge-intensive nature.

\input{tables/overall-fig}

\subsection{Video Segmentation}
In contrast to motion-rich videos, where each frame contributes to the depiction of a continuous movement, presentation recordings are typically composed of static scenes with slides presented by the speaker. 
To reduce the complexity and computational cost,
we segment the video into a sequence of scenes, with one slide representing the visual context of each scene.


Specifically, we first apply the Kernel Temporal Segmentation (KTS) algorithm~\citep{DBLP:conf/eccv/PotapovDHS14,DBLP:conf/iccvw/AfhamSPZSL23} on the video frame features extracted by a pretrained image encoder, e.g., \model{CLIP}~\citep{radford2021clip}. KTS derives the optimal segmentation policy by identifying the boundaries (change points) of each scene from the video frames. The output of KTS is a sequence of consecutive time spans in seconds, with each span indicating the start and end times of a scene.

We find that the predicted scene boundaries align well 
with the moment when the speaker moves to the next slide. Therefore, for each scene, we sample a frame image at 90\% of the time span before the speaker moves to the next slide\footnote{The number of sampled slides from the predicted scenes has an average difference of less than 1.2 pages compared to manually counted number.}. This aims to ensure that the sampled frame contains the complete textual content of the slide even if there were animations.
Processing only 10-30 scene images instead of all video frames significantly reduces computational demands without losing key information.
The scene time spans also guide the segmentation of transcripts corresponding to each slide.
Ultimately, we obtain a sequence of scenes \(\mathcal{S} = [S_1, ..., S_N]\), where \(N\) is the number of scenes. Each scene is represented by a tuple \(S_i = (\mathcal{I}_{i}, \mathcal{T}_{i})\) with the slide image \(\mathcal{I}_{i}\) and the corresponding transcripts \(\mathcal{T}_{i}\), where \(i\) is the scene index.

\subsection{Slide Analysis}

Visual contexts are extracted from the segmented scene images and transformed into natural language.
These textual representations of scenes are further used as
references for transcript refinement.

\paragraph{Visual Context Extraction}

A text-LLM and a vision-LLM collaborate to extract the required visual contexts from the slide images. 
The necessity of using LLMs in both modalities instead of only a vision-LLM lies in the fact that many state-of-the-art vision-LLMs cannot handle complex reasoning tasks, particularly when dealing with false-premise or unanswerable questions~\citep{wang2023evaluation}. (See \Cref{fig:vllm_failure_case} for example)
In contrast, text-LLMs are superior in reasoning and are more robust in these situations. Combining both types of LLMs offers a positive complementary effect.






Specifically, we predefine eight questions covering aspects such as layout, title, bullet points, figures, and tables in the slide that are used to prompt the vision-LLM to extract relevant information
(see questions in Table \ref{tab:predefined_questions} of Appendix \ref{sec:app-implement-detail}).
These answers are often granular, duplicative, excessively lengthy, and even contain misinformation.
Therefore, we regard them as raw contexts, and further leverage a text-LLM to cross-validate, correct inaccuracies caused by the vision-LLM hallucinations, and summarize the lengthy raw contexts into a concise summary, referred to as the scene-level context $\mathcal{C}_{i}$.
This summary is added to the scene tuple $S_i = (\mathcal{I}_{i}, \mathcal{T}_{i}, \mathcal{C}_{i})$. 


\paragraph{Visual Context Condensement}

For certain scenes, information on the current slide might not suffice to correct errors in the corresponding transcripts, as the relevant cues could appear in earlier slides.
Therefore, we use the text-LLM to further condense scene-level summaries into an overall presentation-level summary to share temporal contexts across scenes. This is achieved by prompting the text-LLM to generate a comprehensive summary based on the concatenated scene-level summaries, denoted as $\mathcal{P}$.
Both scene-level and presentation-level summaries will be passed to 
post-edit the initial transcripts.


\subsection{Post-editing}
\label{sec:pe}
In the post-editing stage, we instruct the text-LLM to utilize the extracted visual context $\mathcal{C}_i$ and $\mathcal{P}$ to correct errors in the ASR transcript $\mathcal{T}_i$ (See prompts in Figure \ref{fig:post_editing_prompt} of Appendix \ref{sec:app-implement-detail}). 
The output of the post-editor for each scene is denoted as $\hat{\mathcal{T}}_{i}$. $\sum_{i=1}^{N} \hat{\mathcal{T}}_{i}$ constitutes the final output. 


\paragraph{End-to-end Vision Post-editing}
As \gptfouro has the capability of solving complex vision-language tasks. We set a special step only for \gptfouro to prompt it post-edit the transcript given the slide image in an end-to-end manner.
In this setting slide analysis and post-editing are merged into one step, performed by \gptfouro directly.

%% file: tables/dataset-stats.tex
\begin{table*}[th!]
\centering
\resizebox{1\textwidth}{!}{%
\begin{tabular}{l|cccccc|c}
\hline
\textbf{ID} & \textbf{Gender} & \textbf{L1} & \textbf{Country} & \textbf{Track} & \textbf{Length} & \textbf{\# Slides} & \textbf{DS $\downarrow$} \\ \hline
268 & F & Spanish & Spain & Resources and Evaluation & 00:11:35 & 18 & 0.853 \\
367 & M & Kinyarwanda & Rwanda & Theme: Language Diversity (Best Paper) & 00:11:35 & 18 & 0.824 \\
590 & M & Polish & Poland & Machine Learning for NLP & 00:12:17 & 24 & 0.756 \\
110 & M & - & - & Dialogue and Interactive Systems & 00:12:09 & 27 & 0.683 \\
117 & F & Marathi & India & Question Answering & 00:09:37 & 14 & 0.831 \\
410 & M & Chinese & China & NLP Applications & 00:12:03 & 20 & 0.657 \\
468 & M & - & Belgium & Resources and Evaluation & 00:12:02 & 16 & 0.891 \\
567 & F & Romanian & Romania & Language Grounding, Speech and Multimodality & 00:09:22 & 16 & 0.937 \\
597 & M & Japanese & Japan & NLP Applications & 00:14:02 & 20 & 0.490 \\
111 & M & Hebrew & Israel & NLP Applications & 00:11:53 & 15 & 0.571 \\ \hline
\end{tabular}%
}
\caption{Meta information (video ID, gender, native language (L1), country, track and video length are from \citep{elizabeth2023acl6060}, \#Slides is counted manually) and difficulty score (DS) of the dataset where lower scores indicate higher difficulty.
}
\label{tab:dataset_meta_info}
\end{table*}

%% file: tables/overall-fig.tex
\begin{figure*}[t]
    \centering
    \resizebox{0.8\textwidth}{!}{%
        \includegraphics{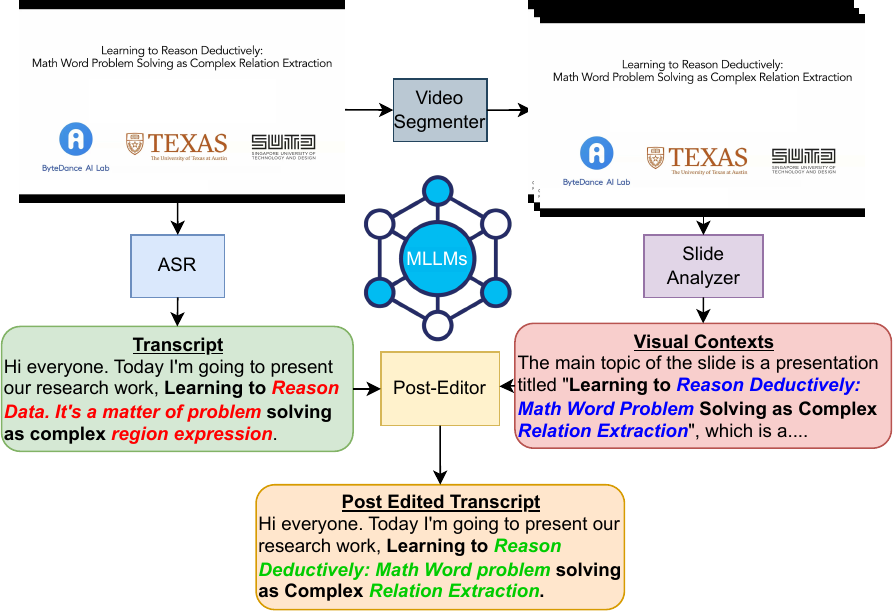}
    }
    \caption{\textbf{The architecture of SciVASR.} \textbf{Baseline ASR} model transcribes a presentation recording into a transcript. \textbf{Video segmenter} splits video frames into a sequence of scenes with each presenting a slide. \textbf{Slide analyzer} applies multimodal LLMs to extract key information from scene images into visual contexts represented by text. \textbf{Post-editor} instruct a text-LLM to leverage visual contexts to refine the ASR transcript.} 
    \label{fig:architecture}
\end{figure*}

%% file: sections/4_experiment.tex
\section{Experiments}

\input{tables/pre-exp-perf-full}
\subsection{Models}
Models used in our experiments involve three types across multiple modalities.

\paragraph{ASR Models} We experiment with \whisperbase and \whisperlarge~\citep{radford2023whisper} as speech recognition models. 
To challenge the effectiveness of our proposed method, we use the inferior \whisperbase as the baseline ASR whose outputs demands more complex refinement. 

\paragraph{Vision-LLMs} We evaluate both open-sourced model \model{CogAgent-VQA}\citep{hong2023cogagent,wang2023cogvlm} given the strengths in GUI and textual-related VQA tasks, and closed-sourced model \gptfouro~\citep{openai2024gpt4}.

\paragraph{Text-LLMs} Based on the observation that models with less than 13B parameters often fail to follow complex instructions and exhibit severe hallucinations, we apply open-sourced \model{Llama3-70B-Instruct}~\citep{llama3modelcard} and closed-sourced \gptfouro\footnote{Model signature: gpt-4o-2024-05-13} in our experiments for effective error correction.

\subsection{Experimental Settings and Metrics}
\label{sec:settings}
We experiment in four settings:
\begin{itemize}
\item{\textbf{ASR-Only}} - an ASR-only baseline using \whisperbase and \whisperlarge.
\item{\textbf{Text-Only Post-editing (Text-PE)}} employs the text-LLM to post-edit the transcripts from the ASR baselines without using any visual information. Post-editing is performed on scene-level transcripts. 
\item{\textbf{Visual-Augmented Post-editing (Vision-PE)}} is the full pipeline of our proposed method presented in Section \S\ref{sec:msr}. It performs post-editing with the visual information. In the open-source model setup, \model{Llama3-70B-Instruct} and \model{CogAgent-VQA} are used.
In the proprietary model setup, \gptfouro deal with both visual information extraction and post-editing.
\item{\textbf{End-to-end Vision Post-editing (E2E-Vision-PE)}} where \gptfouro directly correct the transcript with the slide image without the slide analysis step, as described in \S\ref{sec:pe}.
\end{itemize}

\paragraph{Evaluation Metrics}
Following prior work, we use Word Error Rate (WER) to evaluate the quality of ASR outputs~\citep{elizabeth2023acl6060}. We additionally compute terminology recall given that the dataset provides annotated terminologies in the ground-truth transcript.
%
We also conduct a human evaluation by comparing the raw ASR transcript, the text-only post-edited version, and vision-augmented post-edits. Specifically, the annotators (introduced shortly) are asked to determine which version is better given the ground-truth transcript, without revealing the system they are evaluating. 
Both automatic metrics and human evaluation are conducted at the scene level.\footnote{Original dataset provides the utterance level ground truth, we re-segment both references and hypotheses for each video using \url{https://www-i6.informatik.rwth-aachen.de/web/Software/mwerSegmenter.tar.gz}} 

\paragraph{Annotators}
To guarantee the annotation quality, instead of employing crowd-sourcing annotators, we perform in-house labeling by four annotators who are Master's and PhD students and postdocs and are familiar with ASR.
Two annotators as a group are responsible for 3-4 videos. 
Annotators first independently finish individual annotations and then consolidate their results with the group partner.
In consolidation, partners discuss their disagreements until reaching a consensus.

\subsection{Results and Analysis}
\label{sec:preliminary_result}
The results with four configuration groups are presented in Table~\ref{tab:main_perf_all_setting}.
A negative correlation (-0.636 in spearman-$\rho$) is found between the \texttt{Whisper-base} WER and the difficulty score in \Cref{tab:dataset_meta_info}, indicating that the harder the video, the higher the WER.
In terms of Term-Recall, compared with the baseline \whisperbase, \whisperlarge performs better, which is within our expectations. \textbf{Text-PE} using only the text-LLM for post-editing also improves the term recall, and including visual context (\textbf{Vision-PE}) brings further improvements across all videos. \gptfouro consistently outperforms open-sourced combinations for both text- and vision-PE, the end-to-end setting of \gptfouro obtains comparable performance to its pipelined version.
This exhibits the effectiveness of incorporating visual information. 

\paragraph{Why does post-editing increase WER?}
Despite the significant improvements in term recall, 
WER doesn't reveal consistent improvements.
Open-sourced PE methods even increase the WER in 6 out of 10 videos.
The discrepancy and the counter-intuition motivate us to investigate why the improvements aren't reflected on WER. By analyzing the model outputs, we find that while post-editing accurately 
corrects a certain amount (roughly 40\%) of explicit homophone errors, WER still judges them as errors due to the tiny differences with ground truth text in spelling, e.g., \textit{present} vs. \textit{presents}, and \textit{fine-tuning} vs. \textit{finetuning}.
Additionally, acceptable minor rephrasing caused by the post-editing also increases WER, despite the improvements in fluency. 
These negligible errors are penalized similarly in WER compared with severe errors like recognizing \textit{BERT} as \textit{birds}, making \textbf{WER fail to reflect the true quality of the transcript}, significantly deviating from human judgments as shown in Figure~\ref{fig:winrate}. Post-edited transcripts are consistently voted to be better than the baseline transcript, and \textbf{Vision-PE} is more preferred than \textbf{Text-PE} results.
To address this, we propose a severity-aware ASR evaluation protocol in \S\ref{sec:evaluationframework}.

\input{tables/winrate}

%% file: tables/pre-exp-perf-full.tex
\begin{table*}[ht!]
\centering
\resizebox{1\textwidth}{!}{%
\begin{tabular}{@{}l|rrrr|rrrr|rrrr|rr@{}}
\toprule
 & \multicolumn{4}{c|}{ASR only} & \multicolumn{4}{c|}{Text-PE} & \multicolumn{4}{c|}{Vision-PE} & \multicolumn{2}{c}{E2E-Vision-PE} \\
\multicolumn{1}{c|}{VID} & \multicolumn{2}{c|}{\texttt{Whisper-base}} & \multicolumn{2}{c|}{\texttt{Whisper-large}} & \multicolumn{2}{c|}{Open} & \multicolumn{2}{c|}{\gptfouro} & \multicolumn{2}{c|}{Open} & \multicolumn{2}{c|}{\gptfouro} & \multicolumn{2}{c}{\gptfouro} \\
 & WER$\downarrow$ & \multicolumn{1}{r|}{TR$\uparrow$} & WER$\downarrow$ & TR$\uparrow$ & WER$\downarrow$ & \multicolumn{1}{r|}{TR$\uparrow$} & WER$\downarrow$ & TR$\uparrow$ & WER$\downarrow$ & \multicolumn{1}{r|}{TR$\uparrow$} & WER$\downarrow$ & TR$\uparrow$ & WER$\downarrow$ & TR$\uparrow$ \\ 
 \midrule
110 & 14.41 & \multicolumn{1}{r|}{86.06} & \textcolor{red}{0.35} & \textcolor{optimalgreen}{4.96} & \textcolor{red}{0.70} & \multicolumn{1}{r|}{\textcolor{optimalgreen}{1.82}} & \textcolor{optimalgreen}{-1.23} & \textcolor{optimalgreen}{7.88} & \textcolor{red}{1.17} & \multicolumn{1}{r|}{\textcolor{optimalgreen}{6.06}} & \textbf{\textcolor{optimalgreen}{-2.52}} & \textbf{\textcolor{optimalgreen}{9.70}} & \textcolor{optimalgreen}{-1.41} & \textcolor{optimalgreen}{9.70} \\
117 & 11.16 & \multicolumn{1}{r|}{89.11} & \textcolor{red}{4.31} & \textcolor{optimalgreen}{1.75} & \textcolor{red}{1.84} & \multicolumn{1}{r|}{\textcolor{red}{-0.40}} & \textcolor{optimalgreen}{-1.46} & \textcolor{optimalgreen}{2.02} & \textcolor{red}{5.31} & \multicolumn{1}{r|}{\textcolor{optimalgreen}{0.40}} & \textbf{\textcolor{optimalgreen}{-1.73}} & \textcolor{optimalgreen}{4.84} & \textcolor{optimalgreen}{-0.76} & \textbf{\textcolor{optimalgreen}{5.65}} \\
268 & 18.58 & \multicolumn{1}{r|}{85.92} & \textcolor{red}{5.80} & \textcolor{optimalgreen}{2.96} & \textcolor{red}{1.03} & \multicolumn{1}{r|}{\textcolor{optimalgreen}{4.37}} & \textcolor{optimalgreen}{-1.57} & \textcolor{optimalgreen}{6.80} & \textcolor{red}{1.35} & \multicolumn{1}{r|}{\textcolor{optimalgreen}{7.77}} & \textbf{\textcolor{optimalgreen}{-1.79}} & \textbf{\textcolor{optimalgreen}{11.65}} & \textcolor{red}{0.38} & \textcolor{optimalgreen}{11.65} \\
367 & 17.99 & \multicolumn{1}{r|}{76.44} & \textcolor{optimalgreen}{-1.21} & \textcolor{optimalgreen}{3.89} & \textcolor{optimalgreen}{-2.04} & \multicolumn{1}{r|}{\textcolor{optimalgreen}{4.89}} & \textcolor{optimalgreen}{-3.42} & \textcolor{optimalgreen}{8.44} & \textcolor{optimalgreen}{-1.89} & \multicolumn{1}{r|}{\textcolor{optimalgreen}{9.78}} & \textbf{\textcolor{optimalgreen}{-4.23}} & \textcolor{optimalgreen}{9.33} & \textcolor{optimalgreen}{-3.06} & \textbf{\textcolor{optimalgreen}{10.22}} \\
590 & 14.44 & \multicolumn{1}{r|}{95.73} & \textcolor{red}{5.19} & \textcolor{red}{-2.12} & \textcolor{red}{0.91} & \multicolumn{1}{r|}{\textcolor{optimalgreen}{0.86}} & \textcolor{optimalgreen}{-0.99} & \textcolor{optimalgreen}{0.86} & \textcolor{red}{0.33} & \multicolumn{1}{r|}{\textcolor{optimalgreen}{0.86}} & \textbf{\textcolor{optimalgreen}{-1.07}} & \textbf{\textcolor{optimalgreen}{1.71}} & \textcolor{red}{2.56} & \textcolor{red}{-3.42} \\
$\text{111}^\star$ & 33.12 & \multicolumn{1}{r|}{67.23} & \textcolor{optimalgreen}{-0.87} & \textcolor{optimalgreen}{7.15} & \textcolor{optimalgreen}{-0.89} & \multicolumn{1}{r|}{\textcolor{optimalgreen}{9.60}} & \textcolor{optimalgreen}{-1.13} & \textcolor{optimalgreen}{2.82} & \textcolor{optimalgreen}{-4.13} & \multicolumn{1}{r|}{\textcolor{optimalgreen}{16.95}} & \textbf{\textcolor{optimalgreen}{-5.34}} & \textbf{\textcolor{optimalgreen}{18.08}} & \textcolor{optimalgreen}{-1.05} & \textcolor{optimalgreen}{16.38} \\
$\text{468}^\star$ & 19.81 & \multicolumn{1}{r|}{84.30} & \textcolor{red}{4.32} & \textcolor{optimalgreen}{7.14} & \textcolor{optimalgreen}{-1.35} & \multicolumn{1}{r|}{\textcolor{optimalgreen}{8.14}} & \textcolor{optimalgreen}{-3.40} & \textcolor{optimalgreen}{8.72} & \textcolor{optimalgreen}{-0.18} & \multicolumn{1}{r|}{\textcolor{optimalgreen}{11.63}} & \textbf{\textcolor{optimalgreen}{-5.39}} & \textcolor{optimalgreen}{11.05} & \textcolor{optimalgreen}{-4.28} & \textbf{\textcolor{optimalgreen}{12.79}} \\
$\text{410}^\star$ & 22.99 & \multicolumn{1}{r|}{77.94} & \textbf{\textcolor{optimalgreen}{-2.19}} & \textcolor{optimalgreen}{2.41} & \textcolor{red}{1.72} & \multicolumn{1}{r|}{\textcolor{optimalgreen}{2.21}} & \textcolor{optimalgreen}{-1.55} & \textcolor{optimalgreen}{6.62} & \textcolor{red}{0.92} & \multicolumn{1}{r|}{\textcolor{optimalgreen}{11.77}} & \textcolor{optimalgreen}{-1.49} & \textbf{\textcolor{optimalgreen}{13.97}} & \textcolor{optimalgreen}{-1.26} & \textcolor{optimalgreen}{13.97} \\
$\text{567}^\star$ & 9.73 & \multicolumn{1}{r|}{83.87} & \textcolor{red}{4.90} & \textcolor{red}{-2.82} & \textcolor{red}{1.39} & \multicolumn{1}{r|}{\textcolor{optimalgreen}{1.94}} & \textcolor{optimalgreen}{-0.85} & \textcolor{optimalgreen}{4.52} & \textcolor{red}{0.23} & \multicolumn{1}{r|}{\textcolor{optimalgreen}{7.10}} & \textcolor{optimalgreen}{-0.85} & \textbf{\textcolor{optimalgreen}{10.32}} & \textbf{\textcolor{optimalgreen}{-1.70}} & \textcolor{optimalgreen}{9.68} \\
$\text{597}^\star$ & 28.58 & \multicolumn{1}{r|}{71.84} & \textcolor{red}{0.51} & \textcolor{optimalgreen}{9.17} & \textcolor{optimalgreen}{-1.65} & \multicolumn{1}{r|}{\textcolor{optimalgreen}{6.90}} & \textcolor{optimalgreen}{-3.15} & \textcolor{optimalgreen}{8.05} & \textcolor{optimalgreen}{-5.44} & \multicolumn{1}{r|}{\textcolor{optimalgreen}{16.09}} & \textbf{\textcolor{optimalgreen}{-8.02}} & \textbf{\textcolor{optimalgreen}{18.97}} & \textcolor{optimalgreen}{-4.37} & \textcolor{optimalgreen}{18.97} \\ \midrule
AVG & 19.08 & \multicolumn{1}{r|}{81.85} & \textcolor{red}{2.11} & \textcolor{optimalgreen}{3.45} & \textcolor{red}{0.17} & \multicolumn{1}{r|}{\textcolor{optimalgreen}{4.03}} & \textcolor{optimalgreen}{-1.88} & \textcolor{optimalgreen}{5.67} & \textcolor{optimalgreen}{-0.23} & \multicolumn{1}{r|}{\textcolor{optimalgreen}{8.84}} & \textbf{\textcolor{optimalgreen}{-3.24}} & \textbf{\textcolor{optimalgreen}{10.96}} & \textcolor{optimalgreen}{-1.49} & \textcolor{optimalgreen}{10.56} \\ \bottomrule
\end{tabular}%
}
\caption{This table presents the full settings of our experiment showing the result in WER and Term-Recall (TR). Results are organized into 4 groups with various configurations (introduced in \S\ref{sec:settings}). Videos with $\star$ are from the test set. For clarity, we show the relative improvements compared to the \texttt{Whisper-base} with \textcolor{red}{red} and \textcolor{optimalgreen}{green} distinguishing worse or better performance. Best results are highlighted with \textbf{bold}.}
\label{tab:main_perf_all_setting}
\end{table*}

%% file: tables/winrate.tex
\begin{figure}
    \centering
    \resizebox{1.0\columnwidth}{!}{%
        \includegraphics{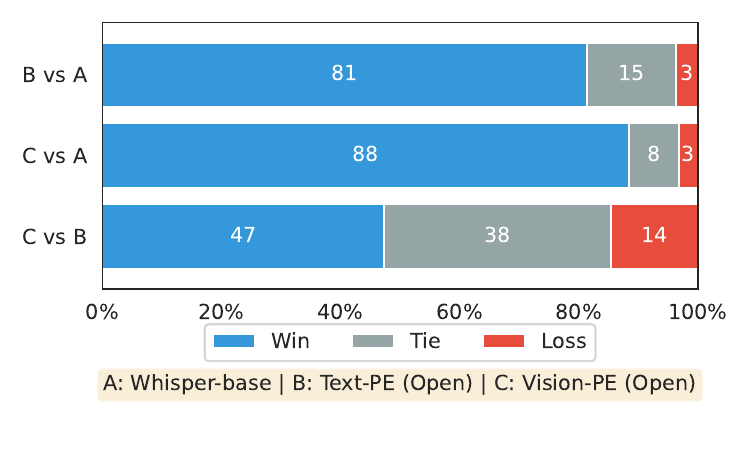}
    }
    \caption{The win, tie and lose ratio between different settings by human annotations, based on \whisperbase and SciVASR in open-source setup.} 
    \label{fig:winrate}
\end{figure}

%% file: sections/5_eval_protocol.tex
\section{Severity-Aware Evaluation Protocol}
\label{sec:evaluationframework}

When evaluating the transcripts, the annotators focus more on the correctness of keywords, terminologies, and numbers, which could impact the meaning of the speech, and care less about minor errors in casing, tense, spelling and grammar. 
This inspires us to propose the severity-aware evaluation protocol, imitating human evaluations.  

\subsection{Breakdown of WER}

WER is computed by counting the number of required operations (insertion, omission, substitution -- $I,O,S$) to transform the hypothesis into the reference and then normalizing it by the reference length $N$. Each operation relates to a mismatch in the hypothesis. 
%
Formally, WER is calculated as 
$\frac{I + O + S}{N}$
where it
assigns equal weights to all operations, without considering the varying importance and the semantic impact of the different types of mismatched content. 
This makes the phenomenon that a transcript with ten errors on general words like ``the'' or ``a'' has the same WER as a transcript with ten errors in terminologies. 
We mitigate it by re-weighting mismatches according to the content type and error severity.

\input{tables/taxonomy}

\subsection{ASR Error Taxonomy}
Inspired by previous studies in NMT that define nuanced error taxonomy and explainable metrics for evaluating the translation hypothesis~\citep{guerreiro2023xcomet,sharou-specia-2022-taxonomy,Freitag_2021}, we propose a taxonomy for ASR errors from two perspectives: content type and severity level of mismatches. The difference between ASR and NMT error taxonomy is discussed in \Cref{sec:related-work}.

\paragraph{What is a mismatch pair?}
Given a pair of (hypothesis, reference), where the hypothesis is the predicted transcript.
A mismatch is a text span differing between hypothesis and reference at the same position, identified by the edit-distance algorithm.

\paragraph{Content Type} 

The content type reflects the nature of the recognized terms in conveying the intended meaning of the spoken language. 
We classify the mismatched content into six types including terminology, numerical data, named entities, grammatical words, disfluencies \& fillers and general words.
We label the content type of each mismatch in the hypothesis by its corresponding span in the reference text -- \textit{ground-truth content}.
Table~\ref{tab:asr_error_taxonomy} shows detailed definitions and corresponding examples. 
Categorizing mismatches by the content type help interpret the nature of each error and capture the full distribution of errors.
\paragraph{Severity Level} 
The content type of a mismatch is not sufficient to determine whether it leads to a severe recognition mistake. 
For example, transcribing the terminology ``finetuning'' as ``fine-tuning'' is tolerable due to a minor formatting change. 
Therefore, we further categorize each mismatch into three levels of severity based on its impact on understanding the text's meaning, including ok, minor and critical, see examples in Table~\ref{tab:asr_error_taxonomy}.
\input{tables/llm-eval-perf}

\subsection{Severity-Aware WER}

\input{tables/tab_reevaluate}

Based on the error taxonomy, we introduce the Severity-aware WER~(SWER) --- WER calibrated by the error severity.
It distinguishes different types of mismatches, making the evaluation of transcripts more accurate and better aligned with human evaluation. Computation of SWER requires three steps.
\paragraph{Pair and highlight mismatches} using the three types of operations for both reference and hypothesis transcripts. We implement this by first using Levenshtein distance to identify the required operations given the transcript reference and hypothesis. Then, three types of parentheses are used for highlighting: ``[]'' for substitutions, ``<>'' for insertions, and ``\{\}'' for omissions. 
The highlighted references and hypotheses are presented in two paragraphs, as the input of the next step: annotating mismatches using the error taxonomy.  

\paragraph{Annotate content type and error severity of mismatch pairs} This can be performed by either human annotators or automatic methods such as LLM-based categorization. We used both in our experiments. 
Note that annotations in this step demand understanding the semantics of the text and making full use of the context. LLMs are recommended leveraging its powerful language understanding capability.

\paragraph{Calculate SWER}
After acquiring the content type and error severity of each mismatch, we can re-weight the importance of each mismatch during the computation of WER. 
We define SWER as
$$\text{SWER} = \frac{\sum_{i=1}^{K} w_i}{N}$$
where $N$ is the reference length, $K$ is the number of mismatches which is same as the total number of operations required in WER, and $w_i$ is the weight corresponding to the severity level of each mismatch. 
In our experiment, we simply define the mapping of severity levels to weights as $w_{critical}$ = 1.0, $w_{minor}$ = 0.6, and $w_{ok}$ = 0.2. The rationale is to down-weight less severe mismatches and place greater emphasis on critical errors. We also tried the setting of (1.0, 0.5, 0.1) and found that (1.0, 0.6, 0.2) gives a better correlation with human evaluations on the dev set.

\subsection{How reliable is LLM evaluation?}


We prompt an LLM to label each mismatch with its content type category and severity level as shown in Figure~\ref{fig:llm_evaluation_case} (see detailed prompt in Figure \ref{fig:llm_evaluation_prompt} in Appendix \ref{sec:app-eval-detail}). \gptfouro{\footnote{Model signature: gpt-4o-2024-05-13} is used in our experiments.

Compared with human annotators who follow the same prompt as guideline to annotate the errors in transcripts produced by ASR-only (\whisperbase results), the LLM performs well in the content type classification task, achieving an F1 score of 93\% as shown in \Cref{tab:llm_evaluator_perf}.
We also convert the severity levels into integers (0,1,2) and compute the Pearson and Spearman correlations between humans and \gptfouro, obtaining significant positive correlations that are more than 0.93. This indicates that \gptfouro is capable of accurately classifying mismatches into the correct categories and assigning reasonable severity levels.

%% file: tables/taxonomy.tex
\begin{table*}[t!]
\centering
\resizebox{1\textwidth}{!}{%
\begin{tabular}{@{}lll@{}}
\toprule
\multicolumn{1}{l|}{\textbf{Names}} & \multicolumn{1}{l|}{\textbf{Definition}} & \textbf{Example} \\ \midrule
\multicolumn{3}{c}{\textsc{Content Type Category}} \\ \midrule \midrule
\multicolumn{1}{l|}{\textbf{Terminology (TERM)}} & \multicolumn{1}{l|}{\begin{tabular}[c]{@{}l@{}}Specialized words or phrases\\ used extensively within a specific field or industry.\end{tabular}} & ``Neural networks'' in a tech discussion \\ \midrule
\multicolumn{1}{l|}{\textbf{Numerical Data (NUM)}} & \multicolumn{1}{l|}{\begin{tabular}[c]{@{}l@{}}Numbers, dates, quantities, or statistical data\\ that provide specific, quantifiable details.\end{tabular}} & ``3.14'', ``thirty percent'' or ``3 out of 4'' \\ \midrule
\multicolumn{1}{l|}{\textbf{Named Entities (NE)}} & \multicolumn{1}{l|}{Names of people, places and organizations.} & ``Albert Einstein'' as a person's name \\ \midrule
\multicolumn{1}{l|}{\textbf{Grammatical Words (GRAM)}} & \multicolumn{1}{l|}{\begin{tabular}[c]{@{}l@{}}Words that primarily serve a grammatical function in a sentence,\\ including articles, prepositions, conjunctions, and auxiliary verbs.\end{tabular}} & ``the'' in ``the cat'' \\ \midrule
\multicolumn{1}{l|}{\textbf{Disfluencies and Fillers (DISF)}} & \multicolumn{1}{l|}{\begin{tabular}[c]{@{}l@{}}Words or phrases that reflect speech disfluency,\\ including fillers, repetitions, or corrections.\end{tabular}} & ``uh'', ``um'', or repeated words due to stuttering \\ \midrule
\multicolumn{1}{l|}{\textbf{General Words (GEN)}} & \multicolumn{1}{l|}{\begin{tabular}[c]{@{}l@{}}Common words that are not specialized terms, named entities,\\ grammatical words, disfluencies, or fillers.\end{tabular}} & ``very'' in ``a very big dog'' \\ \midrule
\multicolumn{3}{c}{\textsc{Severity Level}} \\ \midrule \midrule
\multicolumn{1}{l|}{\textbf{OK (OK)}} & \multicolumn{1}{l|}{No significant impact on the meaning.} & \begin{tabular}[c]{@{}l@{}}REF: ``We try to um try to do''\\ HYP: ``We try to do''\end{tabular} \\ \midrule
\multicolumn{1}{l|}{\textbf{MINOR (MIN)}} & \multicolumn{1}{l|}{Slight impact on the meaning, but the main point is still understandable.} & \begin{tabular}[c]{@{}l@{}}REF: ``this is the number''\\ HYP: ``it is the number"\end{tabular} \\ \midrule
\multicolumn{1}{l|}{\textbf{CRITICAL (CRI)}} & \multicolumn{1}{l|}{Significant or complete change or distortion of the original meaning.} & \begin{tabular}[c]{@{}l@{}}REF: ``finetune BERT''\\ HYP: ``finetune birds''\end{tabular} \\ \bottomrule
\end{tabular}%
}
\caption{The categories of content type and severity level of ASR errors with abbreviations, definitions and examples.}
\label{tab:asr_error_taxonomy}
\end{table*}

%% file: tables/llm-eval-perf.tex
\begin{table}[t!]
\centering
\resizebox{0.8\columnwidth}{!}{%
\begin{tabular}{@{}lrrr@{}}
\toprule
\textbf{Content Type} & \textbf{Precision} & \textbf{Recall} & \textbf{F1} \\ \midrule
\textbf{TERM} & 98.34 & 91.93 & 95.02 \\
\textbf{NUM} & 98.21 & 98.21 & 98.21 \\
\textbf{NE} & 70.00 & 100.00 & 82.35 \\
\textbf{GRAM} & 91.78 & 97.81 & 94.70 \\
\textbf{DISF} & 91.43 & 67.37 & 77.58 \\
\textbf{GEN} & 93.82 & 96.72 & 95.25 \\ \midrule
\textbf{Micro Average} & 94.04 & 93.59 & 93.52 \\ \midrule
\multirow{2}{*}{\textbf{Severity Level}} & \textbf{Pearson} & \multicolumn{2}{c}{\textbf{Spearman}} \\ \cmidrule(l){2-4} 
 & 93.45 & \multicolumn{2}{c}{93.45} \\ \bottomrule
\end{tabular}%
}
\caption{The performance of LLM-based evaluator on \whisperbase transcripts of all videos, against gold labels annotated by humans. Pearson and Spearman correlation coefficients are calculated with three levels of severity converted into integers (0, 1, 2).}
\label{tab:llm_evaluator_perf}
\end{table}

%% file: tables/tab_reevaluate.tex
\begin{table*}[t!]
\centering
\resizebox{1\textwidth}{!}{%
\begin{tabular}{@{}lccccccccc@{}}
\toprule
\multicolumn{1}{l|}{} & \multicolumn{6}{c|}{\textbf{Content-wise Severity Score (weighted/unweighted)$\downarrow$}} & \multicolumn{3}{c}{\textbf{Metrics}} \\ 
\multicolumn{1}{l|}{\textbf{Settings}} & \textbf{TERM} & \textbf{NUM} & \textbf{NE} & \textbf{GRAM} & \textbf{DISF} & \multicolumn{1}{c|}{\textbf{GEN}} & \textbf{WER $\downarrow$} & \textbf{SWER $\downarrow$} & \textbf{Term-Recall $\uparrow$} \\ \midrule
\midrule
\multicolumn{10}{c}{\textit{ASR Baselines}} \\ \midrule
\multicolumn{1}{l|}{\textbf{\whisperbase}} & 271.4/298 & 45.9/111 & 68.2/69 & 42.2/147 & \textbf{14.6/69} & \multicolumn{1}{c|}{349.0/535} & \textbf{19.08} & 5.00 & 81.85 \\
\multicolumn{1}{l|}{\textbf{\whisperlarge}} & \textbf{185.7/237} & \textbf{44.8/110} & \textbf{58.5/66} & \textbf{39.2/117} & 22.6/81 & \multicolumn{1}{c|}{\textbf{336.7/590}} & 21.19 & \textbf{4.62} & \textbf{85.29} \\ \midrule
\midrule
\multicolumn{10}{c}{\textit{SciVASR with Open-sourced LLMs (\model{Llama3-70B-Instruct} + \model{CogAgent-VQA})}} \\ \midrule
\multicolumn{1}{l|}{\textbf{Text-PE}} & 221.6/268 & 42.6/101 & 62.6/65 & \textbf{44.4/176} & 14.6/63 & \multicolumn{1}{c|}{330.8/546} & 19.25 & 4.76 & 85.88 \\
\multicolumn{1}{l|}{\textbf{Vision-PE}} & \textbf{116.0/184} & \textbf{37.6/111} & \textbf{39.2/44} & 44.8/174 & \textbf{13.6/64} & \multicolumn{1}{c|}{\textbf{277.8/585}} & \textbf{18.85} & \textbf{3.44} & \textbf{90.68} \\ \midrule
\midrule
\multicolumn{10}{c}{\textit{SciVASR with Proprietary LLMs (\gptfouro)}} \\ \midrule
\multicolumn{1}{l|}{\textbf{Text-PE}} & 203.2/236 & 41.6/106 & 57.2/63 & 42.4/160 & \textbf{15.4/73} & \multicolumn{1}{c|}{267.4/474} & 17.21 & 4.08 & 87.52 \\
\multicolumn{1}{l|}{\textbf{Vision-PE}} & \textbf{71.2/146} & \textbf{37.0/105} & 18.2/31 & \textbf{39.8/167} & 16.0/72 & \multicolumn{1}{c|}{\textbf{215.0/441}} & \textbf{15.84} & \textbf{2.75} & \textbf{92.81} \\
\multicolumn{1}{l|}{\textbf{E2E-Vision-PE}} & 78.0/160 & 39.8/107 & \textbf{16.4/18} & 43.6/178 & 15.8/71 & \multicolumn{1}{c|}{236.0/484} & 17.59 & 2.81 & 92.40 \\ \bottomrule
\end{tabular}%
}
\caption{Aggregated evaluation results of all settings on both dev and set set. The content-wise severity score with the weighted at left and the unweighted at right for each content type. The weighted score sums up the severity weight of mismatches in this category, and unweighted score is simply the number of mismatches of the category.}
\label{tab:final_perf}
\end{table*}

%% file: sections/6_analysis.tex
\section{Re-Evaluating SciVASR}
\label{sec:reevaluate}
In this section, we re-evaluate our proposed framework using the fine-grained metrics.

\paragraph{Results and Analysis}
As shown in Table \ref{tab:final_perf}, we present metrics including term recall, WER and SWER for overall evaluation. Additionally, we calculate the content-type-wise severity scores for all mismatches in the test set. The weighted score sums up the severity weight of mismatches belonging to this category: $\sum_{i=1}^{K_c} w_{i}$, unweighted score is simply the number of mismatches of the category: $K_c$. 
There are four major findings.

\paragraph{Parameterized knowledge of Text-LLM helps} Compared with ASR baselines, text-only post-editing without visual information reduces both SWER and the content-wise severity scores to some extent.
SWER decreases from 5.0 (\textbf{\whisperbase}) to 4.76  in open-sourced setting and 4.08 in close-sourced setting (\textbf{Text-PE}), indicating that parameterized knowledge in text-LLM is helpful in correcting potential errors.

\paragraph{Visual context is critical for correcting terminology, named entity and numbers} Incorporating visual information brings a significant reduction in the severity scores for TERM (-152), NE (-38) and NUM (-5) in the close-sourced group (\gptfouro \textbf{Vision-PE} vs \textbf{\whisperbase}). SWER reaches 2.75 from 5.0. This suggests that visual modalities are highly effective in correcting these types of errors.

\paragraph{Evaluation with SWER is reliable} SWER can accurately reflect the relationship between the severity of errors and the final scores. This is consistent with the trend presented by the metric of term-recall, a higher correlation of SWER and difficulty score in \Cref{tab:dataset_meta_info} is also obtained for the \texttt{Whisper-base} result (-0.927 in spearman-$\rho$), highlighting the reliability of SWER. In contrast, WER cannot accurately capture this relationship.

\paragraph{Close-sourced setup is superior than the open-sourced one} Though open-sourced PE implementation improves ASR performance, close-sourced \gptfouro still outperforms them, achieving the highest accuracy. However, this is not purely attributed to the powerful capability of \gptfouro, but the pipeline strategy as well, since the \gptfouro end-to-end post-editing setting is inferior to the pipeline method. 
This suggests that despite the strong visual understanding capabilities of \gptfouro, the complexity of this task poses challenges. Our proposed pipeline approach provides an effective solution.

These findings underscore the importance of using visual context to improve ASR performance and validate the efficacy of SWER as a more reliable evaluation metric compared to WER.






%% file: sections/7_conclusion.tex
\section{Conclusion}

This paper introduces the MS-ASR task, addressing the challenge of transcribing knowledge-intensive scientific presentations by integrating visual and speech modalities.
We identified the inefficacy of WER
and proposed the SWER metric to better evaluate ASR performance by considering the content type and error severity.
Through our inference framework -- SciVASR, we demonstrate that state-of-the-art MLLMs, particularly GPT-4o can significantly enhance transcript accuracy by leveraging visual information, improving by 45\% over speech-only baselines.
The experiments underscore the importance of multimodal integration for accurate ASR in complex, real-world scenarios.
Our future works focus on expanding benchmark datasets and refining evaluation protocols to further enhance the applicability of the task.

%% file: sections/limitation_ethical.tex
\section*{Limitation}
We summarize the limitations of our work in the following three aspects:

\paragraph{Benchmark Dataset}
The proposed task would benefit from the inclusion of more datasets as benchmarks, as it currently relies on a single dataset. Expanding the range of datasets will provide a more comprehensive evaluation of the system's performance and enhance its robustness across different domains and types of presentation videos.

\paragraph{Evaluation Protocol}
As the evaluation framework leverages the capabilities of LLMs, it should be used alongside conventional metrics to ensure better reliability. Combining new and traditional evaluation methods will provide a more balanced and accurate assessment of the ASR system's effectiveness, addressing potential biases that might arise from relying solely on LLM-based metrics.

\paragraph{Latency and Error Propagation}
Moreover, the latency and error propagation inherent in our zero-shot framework may pose challenges for practical applications. The time required for processing and the potential for errors to compound through multiple stages can affect real-time usability and overall accuracy. Future work will focus on optimizing these aspects to improve the framework's practicality and reliability.

\section*{Ethical Statement}
In conducting this research, we have adhered to ethical standards to ensure integrity and transparency. All data used in our experiments were obtained from publicly available sources or through collaborations with researchers who provided consent for the use of their materials. We have taken steps to anonymize any potentially sensitive information to protect the privacy of individuals involved in the datasets. Our methods were designed to respect intellectual property rights, and we have appropriately cited all works that have informed our study. Furthermore, our research aims to contribute to the broader scientific community by providing insights and methodologies that can enhance automated speech recognition (ASR) systems, particularly in knowledge-intensive domains. We remain committed to the responsible use of AI technologies, ensuring that our findings are used to benefit educational and scientific advancements while minimizing any potential negative impacts.

%% file: sections/9_acknowledgment.tex
\section*{Acknowledgments}
The authors are grateful to the anonymous reviewers for their helpful comments and corrections.
The first author was supported by an Australian Government Research Training Program (RTP) Scholarship.
This work is supported by the ARC Future Fellowship FT190100039.

%% file: sections/8_appendix.tex
\clearpage
\appendix
\section*{Appendix}
\label{sec:appendix}

\section{Presentation Difficulty Estimation}
\label{sec:difficuty-score}
The difficulty of transcription mainly lies in the accent, fluency, and pronunciation of a speech. 
To eliminate the difficulty stemming from the content, we compare the speech with a standard TTS generation as follows: 
\begin{itemize}
    \item Segment the first 30-second speech chunk from each video and obtain the ground-truth transcript for the chunk. 
    \item Use \textit{OpenAI TTS API} to generate a baseline speech with a standard accent, where the gender of the voice is set as the original speakers'.
    \item Compute the conditional perplexity (CPPL) of the baseline speech (TTS generated) and the real speech by the \whisperbase model, obtaining $\text{CPPL}_{\text{bsl}}$ and $\text{CPPL}_{\text{spch}}$.
    \item Calculate the difficulty score~(DS) as ${\text{CPPL}_{\text{bsl}}/\text{CPPL}_{\text{spch}}}$ reported in Table \ref{tab:dataset_meta_info}. The smaller, the harder.
\end{itemize}

\section{Details about Implementation}
\label{sec:app-implement-detail}

\begin{figure}[ht!]
    \centering
    \resizebox{0.9\columnwidth}{!}{%
        \includegraphics{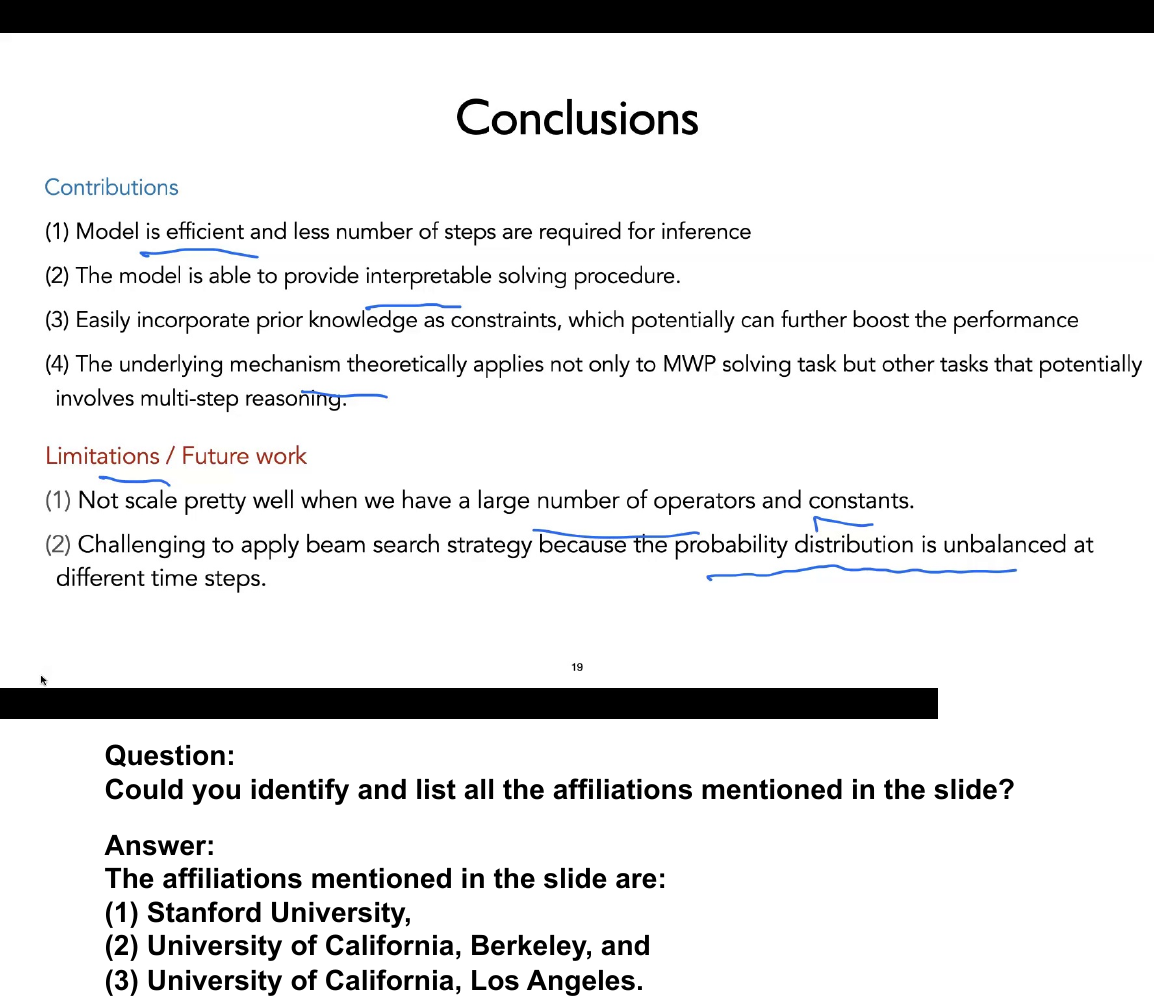}
    }
    \caption{An example showing that vision-LLM could answer with hallucination when the question has false-premise. Image is sampled from video 410, \texttt{CogAgent-VQA} is used for inference.}
    \label{fig:vllm_failure_case}
    \vspace{-1em}
\end{figure}

\Cref{tab:predefined_questions} lists the questions used for visual information extraction. In \Cref{fig:vllm_failure_case}, we show the example of how vision-LLM could hallucinate. 
\begin{table*}[ht!]
\centering
\resizebox{\textwidth}{!}{%
\begin{tabular}{@{}l@{}}
\toprule
\textbf{Questions for vision-LLMs to perform visual-context extraction}                                              \\ \midrule
"Please describe the layout of the slide."                                                                           \\
"What is the main topic or headline of the slide?",                                                                  \\
"Can you list the key points or bullet points presented on the slide?",                                              \\
"Are there any important dates, statistics, or quantitative data mentioned? Please summarize.",                      \\
"Could you identify and list all the affiliations mentioned in the slide?",                                          \\
"Are there any visual aids (e.g., charts, graphs, images, diagrams or tables) on the slide? What information do they convey?", \\
"Does the slide include any quotes, citations, or references to other works? Please detail.",                        \\
"Please provide a holistic summary that encapsulates the entire content and context of the slide.",                  \\ \bottomrule
\end{tabular}%
}
\caption{The predefined questions for vision-LLMs.}
\label{tab:predefined_questions}
\end{table*}

\begin{figure*}[ht!]
    \centering
    \resizebox{1.0\textwidth}{!}{%
        \includegraphics{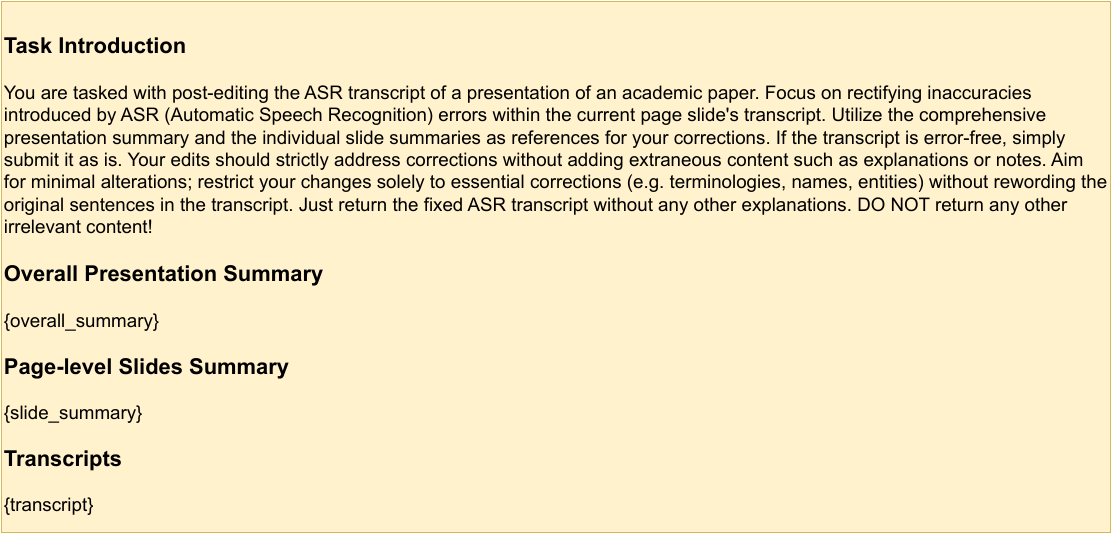}
    }
    \caption{This figure shows the prompt for post-editing with text-LLM.}
    \label{fig:post_editing_prompt}
\end{figure*}

\section{Details about LLM-based Evaluation}
\begin{figure*}[ht!]
    \centering
    \resizebox{1.0\textwidth}{!}{%
        \includegraphics{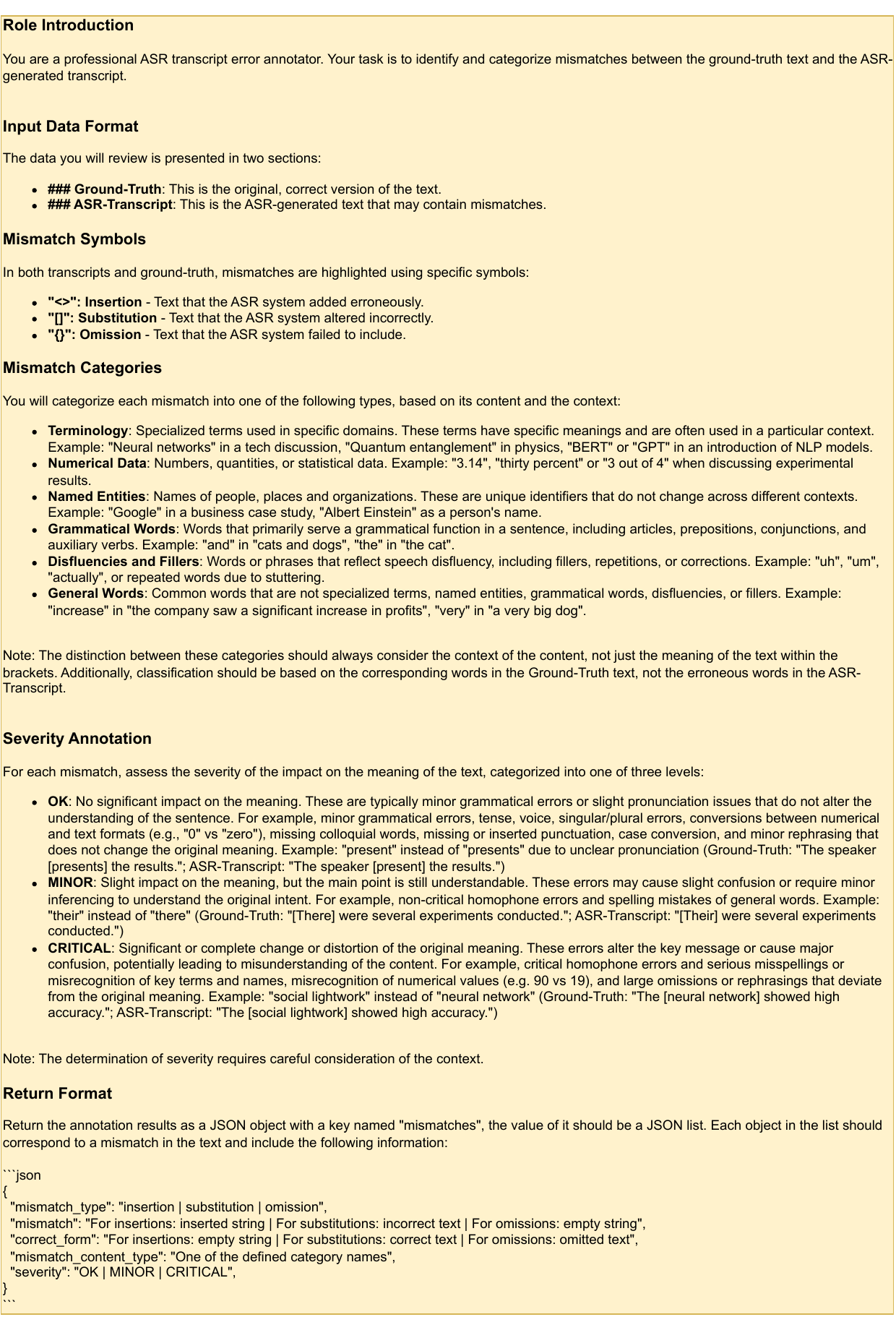}
    }
    \caption{This figure shows the guideline used for automatic evaluation.}
    \label{fig:llm_evaluation_prompt}
\end{figure*}

\begin{figure}[ht!]
    \centering
    \resizebox{0.9\columnwidth}{!}{%
        \includegraphics{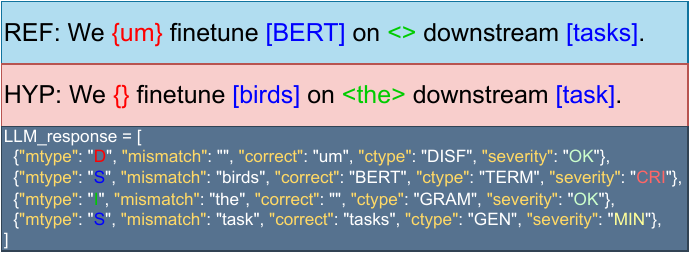}
    }
    \caption{An example of how to leverage LLM to annotate ASR errors based on the taxonomy in Table~\ref{tab:asr_error_taxonomy}. We shorten the mismatch type of omission, substitution and insertion with \textcolor{red}{O}, \textcolor{blue}{S} and \textcolor{green}{I}.}
    \label{fig:llm_evaluation_case}
    \vspace{-1em}
\end{figure}

\Cref{fig:llm_evaluation_prompt} present the annotation guideline for our proposed evaluation protocol, and \Cref{fig:llm_evaluation_case} present the annotated result with the LLM. The performance of LLM-based evaluation is presented in \Cref{tab:llm_evaluator_perf}.
\label{sec:app-eval-detail}

\section{Additional Experimental Results}
We statistic the severity in three configurations in \Cref{tab:severity_statistics}.
\input{tables/severity-statistics}

\input{sections/2_relatedwork}

%% file: tables/severity-statistics.tex
\begin{table}[ht!]
\centering
\resizebox{\columnwidth}{!}{%
\begin{tabular}{@{}lccc|c@{}}
\toprule
\textbf{Severity $\rightarrow$} & \textbf{CRITICAL} & \textbf{MINOR} & \textbf{OK} & \textbf{Total} \\ \midrule
\textbf{\texttt{Whisper-base}} & 529 / 43.04\% & 176 / 14.32\% & 524 / 42.64\% & {\ul \textbf{1229}} \\
\textbf{Text-PE (\gptfouro)} & 328 / 26.69\% & 185 / 15.05\% & 599 / 48.74\% & 1112 \\
\textbf{Vision-PE (\gptfouro)} & 178 / 14.48\% & 181 / 14.73\% & 603 / 49.06\% & 962 \\ \bottomrule
\end{tabular}%
}
\caption{This table presents the absolute error number for each severity level as well as the percentage of it normalized by the baseline error number (underlined).}
\label{tab:severity_statistics}
\end{table}

%% file: sections/2_relatedwork.tex
\section{Related Work}
\label{sec:related-work}
\subsection{Incorporating Visual Modality in ASR}
\citet{oneata2022improving,ghorbani2020listen,gabeur2022avatar,caglayan2019multimodal,wang-etal-2021-make,DBLP:conf/interspeech/WangLGQLSWT0023} propose various approaches to effectively incorporate visual modalities into ASR tasks. These methods typically involve encoding image and speech inputs through visual and audio encoders into dense representations, which are then fused using a modality fusion component such as a transformer encoder or lightweight adapters. This approach ensures that representations of both modalities are mapped into a shared space during training, allowing the decoder to leverage them to predict transcripts in an end-to-end manner during inference.

Despite their success, end-to-end approaches often face significant challenges. One major issue is the scarcity of high-quality triplet data (image, speech, text), which makes it difficult to train models from scratch. To address this, various studies~\citep{jain2024multistage,seo2023avformer} explore leveraging transfer learning on modalities to better utilize unimodal data or introducing an additional stage for self-supervised pretraining to learn the shared representation before fine-tuning the model on the ASR task.


Moreover, existing works are often trained and evaluated on content-less scenarios such as the How2 dataset~\citep{sanabria2018how2}. In these scenarios, vision models only need to provide coarse features, such as object classes, to sufficiently reinforce speech features. However, real-world ASR systems are often used in content-rich scenarios, which are more challenging and require vision models to process more complex content such as text, tables, and figures in slides.

\subsection{Multimodal ASR in the LLM era}
Recent advancements in multimodal large language models (LLMs), such as Video-LLaMA~\citep{zhang-etal-2023-videollama}, Audio-Visual LLM~\citep{shu2023audiovisual}, and AVicuna~\citep{tang2024avicuna}, have demonstrated the ability to leverage both audio and video frames as inputs. However, these models primarily focus on video understanding tasks, paying less attention to speech input. Unlike natural sounds, speech signals often carry more detailed information in a video, making them more challenging for multimodal LLMs to process effectively. While \citet{sun2023finegrained} have explored this area, their work also centers on content-less scenarios such as the How2 dataset~\citep{sanabria2018how2}, leaving the potential of these models in content-rich environments largely unexplored.

\subsection{LLM-based Quality Evaluation}
The reliance on WER for evaluating ASR systems has been increasingly criticized for its inadequacy in reflecting true system performance, mirroring issues in machine translation (MT) where metrics like BLEU fail to capture the full quality spectrum. \citet{sharou-specia-2022-taxonomy} address this by categorizing critical translation errors that could lead to severe user consequences, suggesting a nuanced approach to error analysis. Similarly, \citet{guerreiro2023xcomet} propose a transparent evaluation framework for machine translation, emphasizing fine-grained error detection could be potentially adapted for ASR. \citet{Freitag_2021} also highlight the need for multi-dimensional quality metrics that can provide a more holistic view of translation quality, underscoring the need for ASR evaluations to consider context and practical usability.

The difference between ASR and NMT error taxonomy relies on the interpretation of error, for NMT, errors are often referred to as mistranslation, hallucination or deletions deviate from the source sentence~\citep{sharou-specia-2022-taxonomy}, while ASR errors are referred to mismatches to the ground-truth transcript due to misrecognition. The pursuit of a comprehensive taxonomy of ASR errors is exemplified in \citet{koo2024practical}, they propose an Error Explainable Benchmark (EEB) that categorizes errors at both the speech and text levels, offering a granular understanding of model shortcomings. While this taxonomy is a significant step towards a more detailed diagnosis of ASR systems, its extensive nature may pose challenges in practical implementation due to the complexity of distinguishing between numerous error types.

LLMs have demonstrated their utility in evaluating the quality of generative text, moving beyond surface-level metrics to provide deeper insights into the quality of machine-generated content. \citet{xu2023instructscore} introduces a metric that generates not only a score but also a diagnostic report, identifying specific errors and their severity. Similarly, \citet{fernandes2023devil} presents AUTOMQM, a prompting technique that leverages LLMs to identify and categorize translation errors, aligning with the Multidimensional Quality Metrics (MQM) framework. These approaches underscore the potential of LLMs to offer interpretable and detailed assessments of translation quality, which can be instrumental in refining ASR systems and enhancing user satisfaction.